\documentclass[11pt]{article}

\usepackage{algorithm}               
\usepackage{algpseudocode}           
\usepackage{amssymb,amsmath}         
\usepackage{cite}                    
\usepackage{enumitem}                
\usepackage{epsfig}                  
\usepackage{epstopdf}                
\usepackage[acronym]{glossaries}     
\usepackage{graphicx}                
\usepackage[utf8]{inputenc}          
\usepackage{lmodern}                 
\usepackage[caption=false]{subfig}   
\usepackage{mathtools}

\newcommand{\opram}[1]{$\mathcal{OPRA}_#1$}
\newcommand{\eopram}[1]{$e$\opram{#1}}

\algnewcommand\algorithmicinput{\textbf{Input:}}
\algnewcommand\Input{\item[\algorithmicinput]}

\algnewcommand\algorithmicoutput{\textbf{Output:}}
\algnewcommand\Output{\item[\algorithmicoutput]}

\newcommand{\OPRA}{{\cal OPRA}}

\newacronym{dce}{DCE}{discrete curve evolution}
\newacronym{qsr}{QSR}{qualitative spatial reasoning}
\newacronym{ai}{AI}{artificial intelligence}
\newacronym{qcn}{QCN}{qualitative constraint network}

\newlist{inlinelist}{enumerate*}{1}
\setlist*[inlinelist,1]{label=(\arabic*),}

\title{Towards Applying the OPRA Theory to Shape Similarity}

\author{Christopher H. Dorr, Reinhard Moratz}

\begin{document}

\maketitle

\section{Introduction}
\label{ch:overview}
 
\Gls{qsr} abstracts metrical details of the physical world and enables computers to make predictions about spatial relations even when precise quantitative information is unavailable~\cite{cohn1997qualitative}. From a practical viewpoint \gls{qsr} is an abstraction that summarizes similar quantitative states into one qualitative characterization. A complementary view from the cognitive perspective is that the qualitative method \emph{compares} features within the object domain rather than by \emph{measuring} them in terms of some artificial external scale~\cite{freksa1992using}. As a result, qualitative descriptions are quite natural for humans. For instance, when contemplating the distance to two different destinations, the relative notion ``A is \emph{closer} than B'' is typically more natural than the quantitative alternative, ``A is 160 meters away, and B is 200 meters away.'' Similarly with direction, one may more naturally think of things as ``to the left'' or ``to the right'' instead of in terms of compass bearings or degrees of rotation.

The two main directions in \gls{qsr} are topological reasoning about regions~\cite{randell1992spatial,renz1999complexity,worboys2001integration} and positional reasoning about point configurations, 
like reasoning about orientation and distance~\cite{freksa1992using,clementini1997qualitative,zimmermann1996qualitative,isli1999qualitative,MoratzEtAlAIJ2011}. 
More information on the historical evolution of \gls{qsr} calculi is presented by Moratz in a chapter about
Qualitative Spatial Reasoning in the 
Encyclopedia of GIS
\cite{Moratz2015}.
Additionally, see~\cite{cohn2008qualitative} for an in-depth discussion of the various aspects and approaches in \gls{qsr}.

There is also considerable work about using positional reasoning to describe the qualitative shape of 2D regions~\cite{Schlieder95,schlieder1996qualitative,lovett2010shape}. Many of these approaches represent qualitative shape by listing the relative positions of the adjacent vertices of polygons enumerating the outline of the polygon~\cite{gottfried2002tripartite}. However, these previous approaches make limited use of concepts relating to qualitative distance. Based on recent work by Moratz and Wallgr{\"u}n~\cite{moratz2014spatial}, there is a candidate for a finer resolution positional \gls{qsr} calculus --- \eopram{m}, the ``\textbf{e}levated \textbf{O}riented \textbf{P}oint \textbf{R}elation \textbf{A}lgebra \emph{m}'' --- which utilizes both distance and direction information, and is suited to describe outlines of polygons at different levels of granularity. Both \eopram{m} and its predecessors based around $\OPRA$~(the ``\textbf{O}riented \textbf{P}oint \textbf{R}elation \textbf{A}lgebra'') are discussed in 
detail in \cite{moratz2014spatial}.

The motivation for using qualitative shape descriptions is as follows: qualitative shape descriptions can implicitly act as a schema for measuring the similarity of shapes, which has the potential to be cognitively adequate. Then, shapes which are similar to each other would also be similar for a pattern recognition algorithm. There is substantial work in pattern recognition and computer vision dealing with shape similarity~\cite{prince2012computer}. Here with our
approach to qualitative shape descriptions and shape similarity, the focus is on achieving a representation 
using only simple predicates that a human could even apply without computer support.

To enable verification of a qualitative shape representation by visually comparing shapes with similar descriptions, the representation must be \emph{reversible}. This means it must be possible to take the qualitative shape description and generate prototypical shapes which match the description. In previous work about \gls{qsr}-based shape description, it was only possible to take shapes and generate their \gls{qsr}-based descriptions. It was \emph{not} possible to take a \gls{qsr}-based shape description and let an automatic algorithm generate a sample shape matching the input description. This dissertation presents the first \gls{qsr}-based shape description which intuitively supports the generation of prototypical shapes.

In our Cosit paper \cite{DorrLateckiMoratzCosit2015} we discuss the steps taken to reconstruct simple polygons using their \eopram{m} descriptions. For this task, polygons are defined as a simple closed polylines, or a non self-intersecting chain of line segments in the Cartesian plane $\mathbb{R}^2$. Inputs are converted into qualitative \eopram{m} descriptions, which are then reconstructed as polygons through a combination of state-space searching and constraint propagation.

Roughly, the deconstruction and reconstruction is a three-step process:
\begin{inlinelist}
    \item\label{overview-step1} compute the vertex-pairwise \eopram{m} direction and distance descriptions of the input polyline;
    \item\label{overview-step2} perform an initial reconstruction by ``tracing'' the qualitative hull description;
    \item\label{overview-step3} refine the results of~\ref{overview-step2} via a greedy search.
\end{inlinelist}

Given an appropriate level of granularity, we posit that the \eopram{m} calculus \emph{can} be used to represent and reconstruct similar approximations of simple polygons.
For the technical deatils please refer to our Cosit paper
\cite{DorrLateckiMoratzCosit2015}.
This technical report expands our Cosit paper with results from a computer vision application described in the next section.

\section{High Performance Object Recognition Using Qualitative Shape Similarity}

In a project with cooperation partners Biodiversity Research Institute (BRI), HiDef Aerial Surveying, and SunEdison our research team was sponsored by the U.S. Department of Energy to further develop technology aimed at understanding how birds and bats avoid wind turbines. The collaboration was based around the refinement of a stereo-optic, high-definition camera system developed by HiDef Aerial Surveying. The collaboration deployed systems in order to track flying animals in three dimensions. The technology used two ultra high-definition cameras that are offset, to create a three dimensional view of a wind turbine, the horizon, and an area surrounding the turbine. 

\begin{figure}%

    \centering
    
    \includegraphics[width=0.80\textwidth]{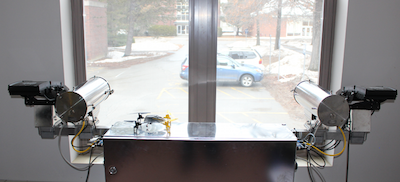}%

    \caption[Stereo-optic camera system with small quadcopter]{\label{fig:camera}%
    Stereo-optic camera system with small quadcopter on top left corner.}%

\end{figure}

Eagles and bats were chosen as the focal species for analyzing camera performance for two reasons: researchers would like to better understand how these species respond to and avoid turbines and both species often receive attention during the permitting process for new wind power projects. Our team from the University of Maine's Robot Interaction Lab worked on algorithms to support partially automated detection of eagles and bats. This is a key component to reduce the analysis time required due to the almost overwhelming amount of data the advanced visual sensors generate.

Developing technology to detect bird and bat avoidance at terrestrial and offshore wind farms will promote a better understanding of the nature of wildlife risks --- or lack thereof --- at wind farms, and reduce uncertainty about the potential for unintended impacts during operation. In the future, these cameras could provide a reliable method of detecting bird and bat response to offshore wind projects, where it is not possible to conduct traditional wildlife monitoring. The goal is to measure how these species behave in the vicinity of a wind farm, how close do they fly, and at what point do they exhibit avoidance behavior. Answering such questions will help wind farms reduce risks to wildlife over the long run.

This project presents a number of unique properties which make it an excellent choice as a testbed for our \gls{qsr} approach. In many real-world computer vision applications, visual data is often highly compressed and noisy --- this is usually a side-effect of sensor quality or processing pipeline limitations. Here, although our cameras operate at an extremely high resolution, there is still the need to compress as much visual information as possible. Further, with the current setup, each camera is monitoring an enormous volume of space --- this means that depending on the distance to the camera, objects of interest vary in pixel dimensions significantly. Combining these two facts leads to visual data which is often incomplete or imperfect. That is to say, there is a often a great deal of uncertainty when dealing with image data from the system, which makes this project an ideal platform for testing our \gls{qsr} shape representations.

Additionally, bird and bat shapes are ideal candidates for a first exploration of our approach, as they possess several unique qualities. Namely, 
\begin{inlinelist}
    \item\label{bird-qualities1} symmetry,
    \item\label{bird-qualities2} simple visual (de)composition of parts,
    \item\label{bird-qualities3} and a natural ``cycle'' of shape states
\end{inlinelist}. 

The first property~\ref{bird-qualities1} ties in nicely with the \eopram{m} notion of rotation invariance, and also enables some level of prediction when dealing with incomplete shapes. The third quality~\ref{bird-qualities3} presents an ideal basis for potential future work on examining the evolution of \gls{qsr} shape representations over time as they relate to a conceptual neighborhood of shape states.

\section{Application: Bird Image Corpus}\label{app:birdcorpus}

In order to test our shape similarity approach, we have worked with researchers at BRI to assemble a corpus of approximately 100 bird images, as captured by the stereo-optic camera system shown in Figure~\ref{fig:camera}. Images in the corpus vary in quality and size, but are mainly of distant birds, and average $\approx~150\times150$~pixels.\footnote{Note that the \emph{actual} screen size of the bird images is usually much smaller -- in cases less than $32\times32$~pixels. Selection of a very small area of interest (AOI) is required to capture many birds at a usable size.}

\begin{figure}
    \centering
    \subfloat[High quality]{{\includegraphics[height=3cm]
    {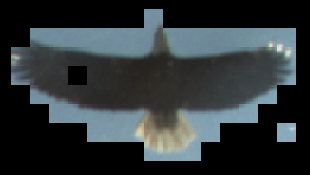}\label{bird:hq} }}\hfill%
    \subfloat[Low quality]{{\includegraphics[height=3cm]
    {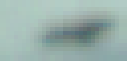}\label{bird:lq} }}\hfill%
    \caption[Corpus image quality]{An example of both high and low quality images from the bird corpus.~\ref{bird:hq} shows an eagle with a native resolution of $\approx256\times128$ pixels, while~\ref{bird:lq} shows an unknown bird with a native resolution of $\approx32\times16$ pixels.}%
    \label{fig:corpusquality}%
\end{figure}

The first step towards applying our shape similarity approach is preprocessing the corpus. The process here differs slightly, but remains similar. Preprocessing is composed of the following steps:

\begin{enumerate}
    \item{Convert the source image to a binary mask}\label{corpuspp:1}
    \item{Extract a single polygon outline from mask}\label{corpuspp:2}
    \item{Simplify the outline with \gls{dce}}\label{corpuspp:3}
\end{enumerate}

\begin{figure}%

    \centering
    
    \includegraphics[width=0.5\textwidth]{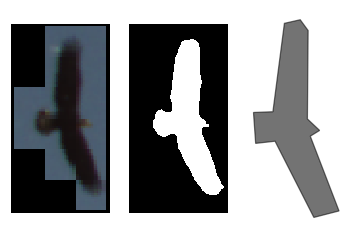}%

    \caption[High quality source image, mask, and outline]{\label{fig:3uphq}%
    High quality source image, mask, and 12-vertex simplified outline. Source image native resolution is $\approx96\times48$ pixels.}%

\end{figure}

\begin{figure}%

    \centering
    
    \includegraphics[width=0.5\textwidth]{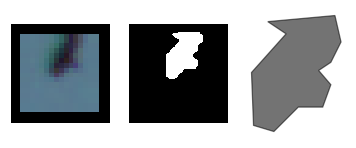}%

    \caption[Low quality source image, mask, and outline]{\label{fig:3uplq}%
    Low quality source image, mask, and 12-vertex simplified outline. Source image native resolution is $\approx16\times16$ pixels.}%

\end{figure}

Given the wide variation in image size and quality, a single fully automatic masking approach was not viable. Instead, Adobe Photoshop was used to perform semi-automatic batch processing on the full corpus of source images. As shown in Figures~\ref{fig:3uphq} and~\ref{fig:3uplq}, mask quality typically depends on source image clarity and completeness. In several instances, the source image does not fully capture a bird, resulting in incomplete or flawed masks.


Regardless of the quality of the mask produced in steps~\ref{corpuspp:1} and~\ref{corpuspp:2}, the next step is to simplify the outlines with \gls{dce}. 
For this application, we have chosen to simplify outlines down to 12 vertices.\footnote{Qualitative shapes will be represented as \eopram{4} objects, where granularity parameter $m=4$, and the number of vertices $=3\times m$.}

\subsection{Corpus Comparison}

With preprocessing complete, the next step in evaluating our shape similarity measure is to begin the process of comparing each unique pair of images from the corpus. Note we are only interested in \emph{unique} pairs; given a comparison function \texttt{cmp(x,y)}: \texttt{cmp(A,B)} == \texttt{cmp(B,A)} (symmetry), and \texttt{cmp(A,A)} = 0 (identity). Given a corpus of $n$ images, this yields $\frac{n^2-n}{2}$ (the number of elements above/below the main diagonal of a square $n\times n$ matrix) unique pairs. For the bird corpus used in this section with $n=97$, we have 4656 unique pairs.

\begin{figure}%

    \centering
    
    \subfloat[Corpus entry 0]
    {\includegraphics[width=0.3\linewidth]%
    {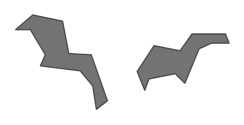}}
    \subfloat[Corpus entry 4]
    {\includegraphics[width=0.3\linewidth]%
    {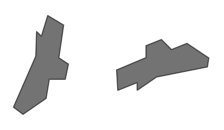}}
    \subfloat[Corpus entry 7]
    {\includegraphics[width=0.3\linewidth]%
    {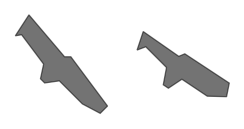}}\\
    \subfloat[Corpus entry 20]
    {\includegraphics[width=0.3\linewidth]%
    {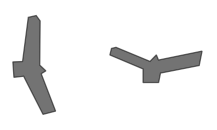}}
    \subfloat[Corpus entry 36]
    {\includegraphics[width=0.3\linewidth]%
    {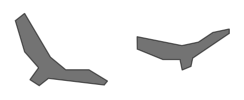}}
    \subfloat[Corpus entry 47]
    {\includegraphics[width=0.3\linewidth]%
    {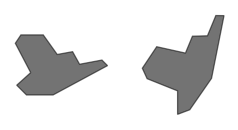}}
    \caption[Several \gls{dce} simplified shapes and reconstructions]{\label{fig:reconstructions}%
    Several \gls{dce} simplified shapes and reconstructions.}%

\end{figure}

To begin the shape comparison, each \gls{dce} simplified outline must first be converted into a qualitative \eopram{4} polygon, as discussed in Section~\ref{generating-qualitative-descriptions}. As the main focus of this application is testing the utility of the \eopram{4} error comparison metrics, the base qualitative shapes are used (instead of their respective reconstructed prototypes). For reference, Figure~\ref{fig:reconstructions} shows several \gls{dce} simplified shapes and their respective reconstructed polygons. Chris Dorr's PhD thesis \cite{DorrThesis} contains a listing of all simplified source shapes and their reconstructions.

Once all simplified source outlines have been converted to qualitative shape representations, the next step is to identify --- for each shape --- what the optimal rotation of every \emph{other} shape is for comparison. The goal of this step is to automatically find the rotation which yields the best alignment between pairs of source outlines. Given 4656 pairs of 12-vertex shapes, this is relatively time consuming\footnote{On a 3.5GHz 6-core Intel Xeon E5, the Python function for finding optimal rotations for each pair in the corpus operates on the scale of 20-30 seconds, while all other steps occur more or less instantly.} and requires testing $4656\times 12\approx56,000$ pairs of rotated shapes. 

After all optimal rotations are discovered, computing the pairwise similarity between \eopram{4} shapes is straight-forward. 
One notable difference between the schema presented previously and the method used here is the addition of a \emph{distance} error measure. Instead of relying only on the qualitative direction comparison, we are now also able to compare the qualitative distances of the edges representing the shape's hull. Initial observations from working with the bird corpus indicated that the distance error measure typically overpowered the direction errors: to this effect direction and distance error measures are computed separately over the entire corpus, and then automatically weighted such that the direction and distance error matrices have the same mean.

As a concrete example, given the corpus of \eopram{4} shapes with 12 vertices, mean direction error over all pairs is $\approx9.26\%$, while mean distance error is $\approx28.37\%$. This gives a ``distance-to-direction error ratio'' (\texttt{dst2dir} ratio) of $\approx3.06$ (mean distance error is roughly three times larger than direction error). To balance the effect of each error input, weights are computed as follows:

$$\textrm{Direction Weight} = \texttt{dst2dir} / (\texttt{dst2dir} + 1)$$
$$\textrm{Distance Weight} = (1 - \textrm{Direction Weight})$$

Applying this to our corpus yields a direction weight of $\approx0.75$, and a distance weight of $\approx0.25$, and scales mean error for both direction and distance to $\approx6.98\%$, effectively balancing their contribution to the total \eopram{4} error.

\subsection{Corpus Comparison Results}

With all qualitative direction and distance errors computed, all that remains is sorting and partitioning the pairs to produce the following sets of results:

\begin{enumerate}
    \item{For each image in the corpus, which other image matches most closely?}\label{corpus:q1}
    \item{For each image in the corpus, which $m$ other images match most closely?}\label{corpus:q2}
    \item{Over the entire corpus, which images are most often matched?}\label{corpus:q3}
\end{enumerate}

\begin{figure}
    \centering

    \subfloat[Shape match with 3.6\% error]%
    {\includegraphics[width=0.6\linewidth]%
    {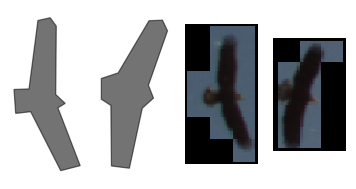}%
    \label{fig:sf1} }

    \subfloat[Shape match with 3.9\% error]%
    {\includegraphics[width=0.6\linewidth]%
    {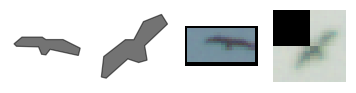}%
    \label{fig:sf2} }



    \subfloat[Shape match with 5.4\% error]%
    {\includegraphics[width=0.6\linewidth]%
    {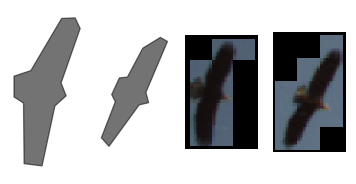}%
    \label{fig:sf5} }
    
    \caption[Top $n=3$ most similar pairs from corpus]{Top $n=3$ closest pairs of images from the bird corpus. Each row shows the simplified shapes as matched, along with their respective source images. Total qualitative error (combined direction and distance) for these pairs ranges from 3.6\% to 5.4\%.}
    \label{fig:pup}
\end{figure}

To answer question~\ref{corpus:q1}, we can simply sort each compared pair by error. Figure~\ref{fig:pup} shows the top three pairs of most similar images from the corpus. 

Note that some of the top matches are representative of sequential images: for example Figure~\ref{fig:sf1} captures the same bird across two frames, and Figure~\ref{fig:sf5} captures the same bird two frames later.

Question~\ref{corpus:q2} can be answered similarly by subsetting the pairs: for each source image, select and sort only other pairs which contain the source image. For a corpus of size $n$, each image will exist in $n-1$ pairs (identity is excluded). Given an arbitrary $m<n-1$, the result is, for each source image, the $m$ most similar other images. Figure 5.7 presents five source images from the corpus, along with their most similar $m=5$ matches. 

\begin{figure}
    \centering

    \subfloat[Corpus image \#20 and top five matches.]%
    {\includegraphics[width=0.7\linewidth]%
    {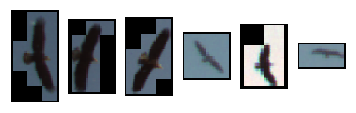}%
    \label{fig:nup1}}

    \subfloat[Corpus image \#21 and top five matches.]%
    {\includegraphics[width=0.7\linewidth]%
    {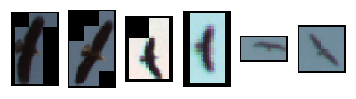}%
    \label{fig:nup2}}

    \subfloat[Corpus image \#22 and top five matches.]%
    {\includegraphics[width=0.7\linewidth]%
    {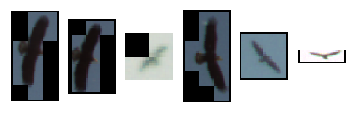}%
    \label{fig:nup3}}

    \subfloat[Corpus image \#69 and top five matches.]%
    {\includegraphics[width=0.7\linewidth]%
    {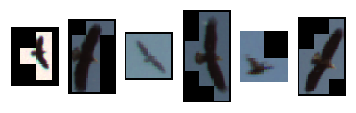}%
    \label{fig:nup4}}

    \subfloat[Corpus image \#91 and top five matches.]%
    {\includegraphics[width=0.7\linewidth]%
    {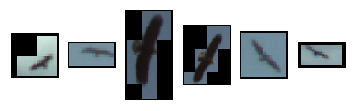}%
    \label{fig:nup5}}
    \caption[Top five most similar five-image sets.]{Five source corpus images with their $m=5$ most similar matching images. For each row, the first (leftmost) image is the base for comparison, while the next five images are the closest matches.}
    \label{fig:nup}
\end{figure}

\begin{figure}
    \centering

    \subfloat[Simplified corpus image \#20 and top five matches.]%
    {\includegraphics[width=0.6\linewidth]%
    {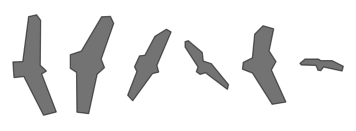}%
    \label{fig:nup1q}}

    \subfloat[Simplified corpus image \#21 and top five matches.]%
    {\includegraphics[width=0.6\linewidth]%
    {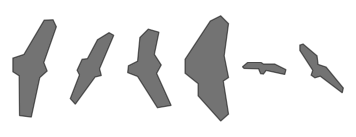}%
    \label{fig:nup2q}}

    \subfloat[Simplified corpus image \#22 and top five matches.]%
    {\includegraphics[width=0.6\linewidth]%
    {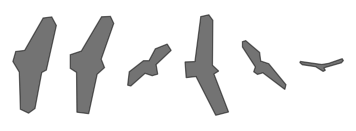}%
    \label{fig:nup3q}}

    \subfloat[Simplified corpus image \#69 and top five matches.]%
    {\includegraphics[width=0.6\linewidth]%
    {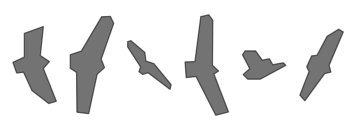}%
    \label{fig:nup4q}}

    \subfloat[Simplified corpus image \#91 and top five matches.]%
    {\includegraphics[width=0.6\linewidth]%
    {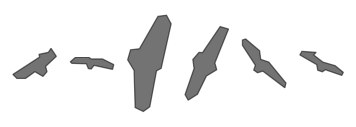}%
    \label{fig:nup5q}}
    \caption[Simplified shapes for top five most similar five-image sets.]{\gls{dce} simplified outlines of the image sets from Figure~\ref{fig:nup}.}
    \label{fig:nupq}
\end{figure}

\begin{figure}
    \centering

    \subfloat[Corpus image \#49 and top five matches.]%
    {\includegraphics[width=0.6\linewidth]%
    {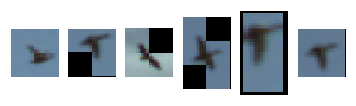}%
    \label{fig:ducks-src} }

    \subfloat[\gls{dce} simplified outlines of source, top five matches.]%
    {\includegraphics[width=0.6\linewidth]%
    {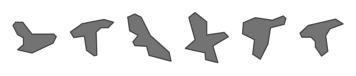}%
    \label{fig:ducks-qpl} }
    
    \caption[Set of $m=5$ closest matches for a duck.]{Top $m=5$ matches for corpus image \#49. Mean error between all shapes in the set is $\approx 10\%$.}
    \label{fig:nupducks}
\end{figure}

In Figures~\ref{fig:nup} and~\ref{fig:nupq}, mean error between all images/shapes in a row is within 6-7\%. As with results from question~\ref{corpus:q1}, some of the top matches for a given image are from a sequence: particularly the source images for \ref{fig:nup1}, \ref{fig:nup2}, and \ref{fig:nup3}, which are from a three-frame sequence. While the majority of the images in the corpus are of eagles or gulls, there are also a number of more distinctly shaped birds, such as the set of ducks shown in Figure~\ref{fig:nupducks}.

\begin{figure}
    \centering

    \subfloat[Top five most-matched corpus images.]%
    {\includegraphics[width=0.6\linewidth]%
    {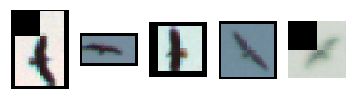}%
    \label{fig:top5-src} }

    \subfloat[\gls{dce} simplified outlines of top five most-matched images.]%
    {\includegraphics[width=0.6\linewidth]%
    {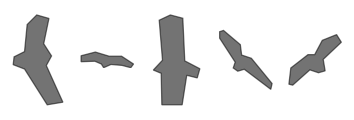}%
    \label{fig:top5-qpl} }
    
    \caption[Five most frequently matched images]{Five most frequently matched images and respective \gls{dce} simplified outlines when looking at top $m=5$ similar images over entire corpus.}
    \label{fig:top5}
\end{figure}

Lastly, to address question~\ref{corpus:q3}, we can simply tally, for each image in the corpus, how many times it appears as a match to other images. Figure~\ref{fig:top5} shows the five most commonly matched images when looking at sets of the $m=5$ nearest matches. Across 96 5-image sets, the images in Figure~\ref{fig:top5} were matched between 12 and 17 times. 

Although the source image quality is somewhat low, the \gls{dce} simplified outlines from Figure~\ref{fig:top5} each represent a fairly generic bird pose, which combined with the relatively low quality of most corpus images, likely contributes to their recurring inclusion across multiple sets.

The PhD thesis \cite{DorrThesis} lists all 96 samples and contains the complete results for questions~\ref{corpus:q1} and~\ref{corpus:q2}.




\clearpage

\bibliographystyle{plain}

\bibliography{thesis,thesis2,epra1,epra2,latecki,misc,tpcc}

\end{document}